\pdfoutput=1

\documentclass[11pt]{article}

\usepackage[preprint]{acl}

\usepackage{times}
\usepackage{latexsym}

\usepackage[T1]{fontenc}

\usepackage[utf8]{inputenc}

\usepackage{microtype}

\usepackage{inconsolata}

\usepackage{graphicx}

\title{DB-Explore: Automated Database Exploration and Instruction Synthesis for Text-to-SQL}

 \author{
 \textbf{Haoyuan Ma}\textsuperscript{1}, 
 \textbf{Yongliang Shen}\textsuperscript{1}\textsuperscript{\textdagger}, 
 \textbf{Hengwei Liu}\textsuperscript{1},
 \textbf{Wenqi Zhang}\textsuperscript{1},
 \textbf{Haolei Xu}\textsuperscript{1}, \\
 \textbf{Qiuying Peng}\textsuperscript{2}
 \textbf{Jun Wang}\textsuperscript{2}, 
 \textbf{Weiming Lu}\textsuperscript{1}\textsuperscript{\textdagger}  
 \\
$^{1}$Zhejiang University,
$^{2}$ OPPO Research Institute \\
\texttt{\{mahaoyuan, syl, luwm\}@zju.edu.cn} \\
\texttt{\{pengqiuying, wangjun7\}@oppo.com}  
}

\begin{document}
\maketitle

\renewcommand{\thefootnote}{\textdagger}
\footnotetext{Corresponding author.}

\begin{abstract}
Recent text-to-SQL systems powered by large language models (LLMs) have demonstrated remarkable performance in translating natural language queries into SQL.
However, these systems often struggle with complex database structures and domain-specific queries, as they primarily focus on enhancing logical reasoning and SQL syntax while overlooking the critical need for comprehensive database understanding.
To address this limitation, we propose DB-Explore, a novel framework that systematically aligns LLMs with database knowledge through automated exploration and instruction synthesis.
DB-Explore constructs database graphs to capture complex relational schemas, leverages GPT-4 to systematically mine structural patterns and semantic knowledge, and synthesizes instructions to distill this knowledge for efficient fine-tuning of LLMs.
Our framework enables comprehensive database understanding through diverse sampling strategies and automated instruction generation, bridging the gap between database structures and language models.
Experiments conducted on the SPIDER and BIRD benchmarks validate the effectiveness of DB-Explore, achieving an execution accuracy of 67.0\% on BIRD and 87.8\% on SPIDER. 
Notably, our open‑source implementation based on Qwen2.5‑Coder‑7B achieves state‑of‑the‑art results at minimal computational cost, outperforming several GPT‑4‑driven Text‑to‑SQL systems. 
\end{abstract}

\section{Introduction}
Text-to-SQL systems have emerged as a crucial bridge between humans and structured data, enabling natural language interaction with databases without requiring SQL expertise. While recent advances in large language models (LLMs), including ChatGPT \citep{liuComprehensiveEvaluationChatGPTs2023a}, GPT-4 \citep{achiam2023gpt}, and Qwen2 \citep{hui2024qwen2}, have significantly improved these systems' capabilities, demonstrating enhanced comprehension of user queries and the ability to generate complex SQL queries, they still face fundamental challenges in understanding complex database structures and generating accurate SQL queries, particularly for domain-specific applications \citep{zhangSurveyTableReasoning2024a, liuComprehensiveEvaluationChatGPTs2023a}.

A key limitation of current approaches is their reliance on direct query translation without deep understanding of the underlying database structure and semantics. Existing methods primarily focus improving prompting strategies \citep{dongC3ZeroshotTexttoSQL2023, rajkumarEvaluatingTexttoSQLCapabilities2022, liuComprehensiveEvaluationChatGPTs2023a, sql-plam} or enhancing the reasoning ability of LLMs through supervised fine-tuning \citep{liRESDSQLDecouplingSchema2023, scholakPICARDParsingIncrementally2021, wangRATSQLRelationAwareSchema2020}, but often overlook the critical role of database comprehension in query generation. While techniques like schema linking \citep{wangRATSQLRelationAwareSchema2020, liRESDSQLDecouplingSchema2023}, question decomposition \citep{pourrezaDINSQLDecomposedInContext2023} and self-correction \citep{wangMACSQLMultiAgentCollaborative2023} have attempted to address this gap, they typically treat the database as a static structure rather than a rich source of semantic knowledge that can inform the query generation process.

We argue that effective Text-to-SQL systems must first develop a comprehensive understanding of the database through systematic exploration before attempting query generation. This exploration should capture not only the structural relationships between tables but also the semantic patterns and domain-specific knowledge embedded within the schema. However, automated database exploration presents several challenges:
(1) \textit{Complex structural relationships:} Real-world databases often contain intricate networks of table relationships that are difficult to capture and represent effectively.
(2) \textit{Hidden semantic patterns:} Domain-specific knowledge embedded in schema elements (table names, column names, etc.) is often implicit and requires careful analysis to uncover.
(3) \textit{Query complexity progression:} Understanding how simple queries can be systematically extended to handle more complex scenarios requires a structured approach to knowledge acquisition.

To address these challenges, we propose DB-Explore, a novel framework that transforms Text-to-SQL training through systematic database exploration and instruction synthesis. At the core of our approach is a graph-based database representation technique called DB Graph, which captures both structural relationships and semantic patterns within the database. This representation serves as a foundation for three key components:
(1) \textit{Semantic Knowledge Extraction:} We develop techniques to uncover domain-specific knowledge embedded in schema elements, enabling more contextually appropriate query generation.
(2) \textit{Structural Pattern Mining:} By analyzing the DB Graph, our system identifies common join patterns, table relationships, and query templates that reflect the database's architectural design.
(3) \textit{Progressive Instruction Synthesis:} Using the extracted knowledge, we automatically generate training examples that progress from simple to complex queries, allowing models to gradually master increasingly sophisticated query patterns.


Our framework leverages these components to create comprehensive training datasets that capture both the structural and semantic complexity of the target database. This approach fundamentally differs from existing methods by treating database exploration as a crucial prerequisite for effective query generation, rather than relying solely on input-output pairs or schema information.
The main contributions of this work are:

\begin{enumerate}
\item We introduce DB-Explore, a framework that revolutionizes Text-to-SQL training by incorporating systematic database exploration and knowledge extraction.
\item We develop DB Graph, a novel representation that captures both structural relationships and semantic patterns within databases, enabling more effective exploration and learning.
\item We present techniques for automated instruction synthesis that leverage extracted database knowledge to create progressively complex training examples.
\item We demonstrate significant improvements in query generation accuracy and complexity handling across multiple benchmark datasets with minimal computational cost.
\end{enumerate}

\section{Related Work}

\begin{figure*}[ht]
  \includegraphics[width=2\columnwidth]{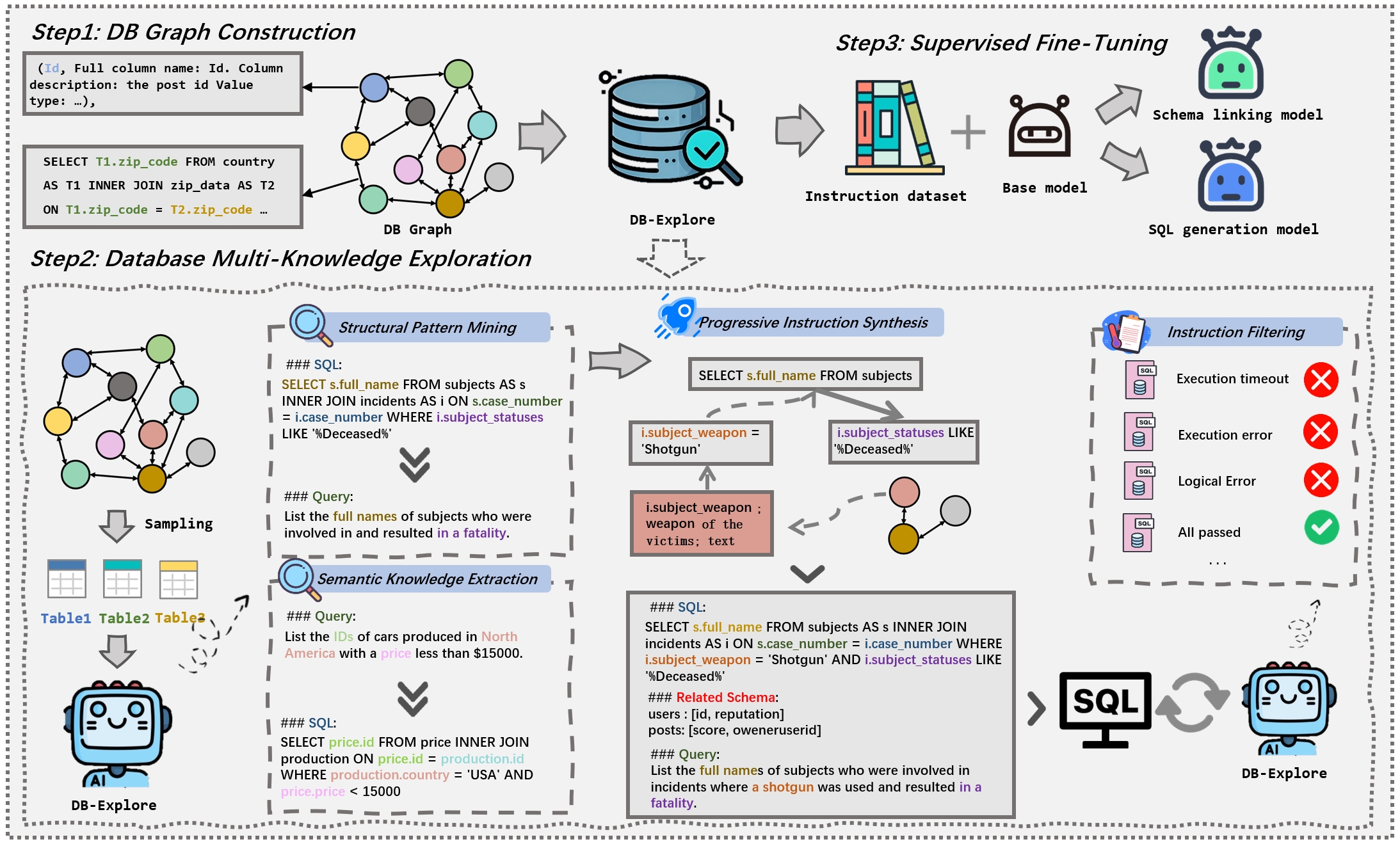}
  \caption{\label{fig1}Overall framework of DB-Explore. Our framework operates through three principal phases: (1) Database Graph Construction, (2) Database Multi-Knowledge Exploration: semantic knowledge extraction, structural pattern mining, progressive instruction synthesis, and filtering mechanisms, and (3) Supervised Fine-Tuning executing database-adaptive model training. The synthesis phase systematically produces database-consistent training corpora through multi-stratum knowledge alignment.}
\end{figure*}

Early Text-to-SQL systems relied on handcrafted, rule-based parsers to map natural language into SQL \citep{li2014constructing, popescu2003towards}. The advent of large language models (LLMs) such as GPT-4 \citep{achiam2023gpt} shifted focus to leveraging rich language priors for SQL generation \citep{liCanLLMAlready2023}. Building on this paradigm, schema-focused methods such as RESD-SQL, DTS-SQL, and RSL-SQL enhance schema linking and selection to ground queries in database structures \citep{liRESDSQLDecouplingSchema2023, pourrezaDTSSQLDecomposedTexttoSQL2024, caoRSLSQLRobustSchema2024b}, while decomposition-based techniques like DIN-SQL, PTD-SQL, and QPL break complex queries into simpler sub‑tasks \citep{pourrezaDINSQLDecomposedInContext2023, luo2024ptd, qpl}. MSC-SQL, SQL-PaLM, Distillery, and XiYan improve robustness through multi‑candidate generation and ranking \citep{gortiMScSQLMultiSampleCritiquing2024, sql-plam, maamariDeathSchemaLinking2024a, gao2024xiyan}, and skeleton‑aware selection in DAIL-SQL refines few‑shot demonstrations \citep{dail}. Self‑correction frameworks in MAC and EPI-SQL further reduce errors via iterative feedback \citep{wangMACSQLMultiAgentCollaborative2023, liu2024epi}.

In parallel, competitive open‑source LLMs have spurred fine‑tuning frameworks that exploit synthetic or task‑specific data. CODES incrementally pre‑trains StarCoder on SQL corpora and integrates external plugins for enhanced performance \citep{liCodeSBuildingOpensource2024}, whereas DTS-SQL and SENSE employ schema‑aware fine‑tuning and direct preference optimization (DPO) \citep{rafailov2024direct} to improve accuracy \citep{pourrezaDTSSQLDecomposedTexttoSQL2024, sense}. ROUTE unifies multi‑task tuning through targeted data synthesis \citep{ROUTE}. Recently, reinforcement learning (RL) methods have emerged as a promising direction in the Text-to-SQL field. SQL-o1 \citep{sql-o1} employs Monte Carlo Tree Search (MCTS) \citep{monte} to enhance the reasoning process during SQL generation. SQL-R1 \citep{sql-r1} and Reasoning-SQL \citep{reasoning-sql} leverage the GRPO \citep{deepseekv3} framework and incorporate domain-specific reward functions, improving the accuracy and robustness of Text-to-SQL systems. Despite these advances, most work remains focused on direct SQL generation, often overlooking holistic database understanding.


\section{Methodology}


In this section, we present a detailed explanation of DB-Explore, a fully automated framework that leverages LLMs to explore databases, generate data, and perform supervised fine-tuning. As illustrated in Figure \ref{fig1}, the framework comprises three core components: 1) Database Graph Construction, 2) Database Multi-Knowledge Exploration, and 3) Supervised Fine-Tuning.

DB-Explore initiates with the construction of DB Graph, which captures the intricate topological relationships within the database. By applying various sampling strategies, we extract subgraphs containing diverse information. DB-Explore then converts this database knowledge into a instruction dataset through semantic knowledge extraction, structural pattern mining, and progressive instruction synthesis. To ensure the quality of the generated instructions, an instruction filtering mechanism is employed. Finally, the extracted knowledge is injected into the base model via supervised fine-tuning. In the following subsections, we present a detailed description of each component of DB-Explore. 

\subsection{Database Graph Construction and Sampling}

To effectively model both structural and semantic relationships within databases, we formalize database schemas as DB graph, as illustrated in Step 1 of Figure \ref{fig1}. The graph comprises nodes representing database columns and edges encoding relational constraints. Two edge types are defined: Intra-table edges connect columns within the same table, enabling direct access without JOIN operations; Inter-table edges represent foreign key relationships across tables, bridging subgraphs into a unified structure. Edge weights are assigned based on co-occurrence frequency of connected columns in seed queries, prioritizing frequently accessed schema patterns that reflect real-world query behaviors.

To enable more effective exploration and learning, we propose two complementary sampling strategies: 
1) \textbf{Random Walk Sampling}: Starting from selected nodes, the walker iteratively traverses adjacent nodes via randomly chosen edges until reaching a predefined sampling size. This method ensures comprehensive schema coverage but may overlook high-priority relationships.
2) \textbf{Empirical Weighted Sampling}: Normalized edge weights guide traversal probabilities, biasing selection toward high-weight edges. This strategy emphasizes schema subgraphs aligned with user query patterns, enhancing instruction relevance.
Both strategies generate subgraphs as input for subsequent database multi-knowledge exploration, balancing semantic diversity via random walks and user intent alignment via weighted sampling.

\subsection{Database Multi-Knowledge Exploration}
We propose a three-pronged exploration and data synthesis framework to explore database by instruction synthesis while minimizing computational overhead: 1) Semantic Knowledge Extraction, 2) Structural Pattern Mining, and 3) Progressive Instruction Synthesis. 

\subsubsection{Semantic Knowledge Extraction}

Semantic knowledge extraction mechanism leverages LLMs' internal knowledge to discover latent schema relationships through a self-guided paradigm. The process initiates by sampling schema subgraphs $S$ from the DB Graph via random walk sampling, containing table/column metadata and structural relationships.

Self-instruction has been shown to be an effective way to synthesize data \citep{wang2022self}. The instruction synthesis pipeline operates through a self-instruct framework initialized using manually-crafted template.  We prompt the LLM with $S$ and randomly sampled seed instructions as demonstrations to synthesize novel queries $Q$ through schema-constrained synthesis. The generated ⟨$Q$, $S$⟩ pairs are subsequently feedback to the LLM to produce executable SQL responses $R$.

To achieve optimal semantic diversity while preserving logical validity, we implement temperature scaling ($\tau$=0.8) during generation. The iterative process progressively enriches the instruction pool through random sampling, thereby expanding coverage of potential user intents.

\subsubsection{Structural Pattern Mining}

Unlike semantic knowledge extraction, structural pattern mining is designed to generate schema-grounded SQL queries through explicit modeling of relational topologies derived from the DB Graph. It prioritizes foreign-key dependencies and latent connectivity patterns observed in seed instructions, while strategically balancing single-table operations with multi-tables queries through weighted edge sampling. We propose a SQL-first generation paradigm to ensure structural accuracy by initially producing executable SQL statements that rigorously adhere to schema constraints. These syntactically valid queries are subsequently translated into natural language instructions via inverse parsing, effectively reducing hallucination risks compared to open-ended synthesis.

As shown in Figure~\ref{fig1}, we first apply empirical weighted sampling to select schema subgraphs $S$, giving priority to high-frequency foreign-key edges. For each selected subgraph, SQL statements $R$ are automatically generated by traversing all nodes in the subgraph, incorporating JOIN operations based on foreign keys, filter conditions derived from column data types, and aggregation functions applied to numerical columns. The resulting SQL statements are then translated into natural language queries using the back-instruct method \citep{shenTaskBenchBenchmarkingLarge2023}, where an LLM rewrites each SQL query $R$ into a structurally faithful natural language instruction $Q$.


\subsubsection{Progressive Instruction Synthesis}

In this section, we introduce a progressive instruction synthesis mechanism that progressively enriches query complexity to address the challenge of generating complex queries from simple user inputs. As illustrated in Figure \ref{fig1}, SQL queries are formalized as syntax trees where the root represents the core SELECT statement and leaf nodes correspond to schema-specific constraints. Each leaf node maps to a modular condition that can be independently translated into natural language, establishing a bidirectional bridge between SQL semantics and user intent.  

The synthesis process initiates with a base query sampled from the DB Graph. Through iterative refinement:  
1. Condition Augmentation: Schema-derived nodes (foreign keys, constraints) are recursively appended to the syntax tree via DB Graph-guided sampling. 
2. Complexity Scaling: The LLM orchestrates tree expansion by injecting compound conditions such as nested WHERE clauses and multi-table JOIN while preserving syntactic validity. 
3. Instruction Rewriting: Each evolved SQL tree is verbalized into diverse natural language expressions through LLM rewriting. 

Each iteration introduces new constraints and conditions to the instruction, progressively increasing its complexity. We set the number of iterations to 3, corresponding to user instructions of varying difficulty levels: simple, moderate, and challenging. This progressive paradigm generates instruction pairs ranging from atomic queries to multi-hop reasoning tasks, effectively simulating the evolutionary patterns of real-world user queries.


\subsection{Instruction Filtering}
Progressive instruction synthesis often generates a significant amount of noisy data due to hallucinations of LLMs. To mitigate noise data, we implement a dual-stage validation pipeline:

All generated SQL candidates $R$ are executed against the target database, and queries are discarded if they contain syntax errors, reference invalid columns or tables, or exceed an execution time limit of 25 seconds. The surviving $⟨Q,R⟩$ pairs are then subjected to LLM-based consistency verification to ensure faithful alignment between the natural language intent and the SQL logic, as well as strict adherence to schema constraints defined in $S$. Validated instances are formatted as augmented tuples $⟨Q,D,R,S⟩$, where $D$ denotes the database context. These high-quality instances are reintegrated into the seed dataset, forming a self-improving feedback loop that drives iterative synthesis and database exploration. 



\subsection{Supervised Fine-Tuning and Inference} 
To inject the database knowledge into base model, we implement a two-stage supervised fine-tuning framework using the synthesized instruction dataset $⟨Q, D, R, S⟩$. Addressing the challenge of schema complexity versus limited context windows, user queries undergo schema linking to extract critical schema elements $S$ before SQL generation. This cascaded approach reduces irrelevant schema noise and minimizes token consumption. 

To enhance the model's comprehension of database structures during inference, we integrates a Value Retrieval Module which implements a coarse-to-fine matching mechanism via BM25 indexing and Longest Common Subsequence (LCS) algorithms \citep{liCodeSBuildingOpensource2024}, resolving token mismatches. Additionally, metadata augmentation is utilized, embedding database descriptions, column types, foreign keys, and domain knowledge into prompts. 



\section{Experiments}

\begin{table*}[ht]
  \centering
  \begin{tabular}{lcccc}
    \hline
    Methods                       & \multicolumn{2}{c}{SPIDER}    & \multicolumn{2}{c}{BIRD} \\
    ~                               & Dev-EX     & Dev-TS      & Dev-EX      & Dev-VES \\
    \hline
    \multicolumn{5}{l}{\textit{Prompting with Closed-Source LLMs}} \\
    MCS-SQL + GPT-4 \citep{mcs}& 89.5          & -              & 63.4           & 64.8    \\
    XIYAN \citep{gao2024xiyan}                         & 89.7          & -              & 73.3           & -    \\
    CHASE-SQL + Gemini-1.5 \citep{chase}        & 87.6          & -              & 73.1           & -    \\
    GPT-4 \citep{achiam2023gpt}                         & 72.9          & 64.9           & 46.4           & 49.8    \\
    DIN-SQL + GPT-4 \citep{pourrezaDINSQLDecomposedInContext2023}               & 82.8          & 74.2           & 50.7           & 58.8    \\
    DAIL-SQL + GPT-4 \citep{dail}              & 83.5          & 76.2           & 54.8           & 56.1    \\
    MAC-SQL + GPT-4 \citep{wangMACSQLMultiAgentCollaborative2023}               & 86.8          & 82.8           & 59.4           & 66.2    \\
    TA-SQL + GPT-4 \citep{tasql}                & 85.0          & -              & 56.2           & -    \\
    MAG-SQL + GPT-4 \citep{mag}               & 85.3          & -              & 61.1           & -    \\
    \hline
    \multicolumn{5}{l}{\textit{Fine-Tuning with Open-Source LLMs 7B}} \\
    DTS-SQL + DeepSeek-7B \citep{pourrezaDTSSQLDecomposedTexttoSQL2024}                    & 82.7$^*$          & 78.4$^*$           & 55.8           & 60.3\\
    CODES-7B \citep{liCodeSBuildingOpensource2024}                     & 85.4& 80.3& 57.2& 58.8\\
    SENSE-7B \citep{sense}                     & 83.2          & \underline{81.7}           & 51.8           & -    \\
    ROUTE + Qwen2.5-7B \citep{ROUTE}            & 83.6          & 77.5           & 55.9& 57.4    \\
    OmniSQL-7B \citep{omnisql}            & 81.6          & -           & 63.9& -    \\
    SQL-o1 + Qwen2.5-7B \citep{sql-o1}            & 84.7          & 78.5           & \underline{66.7} & \underline{70.4}   \\
    SQL-R1(self-consistency@8) \citep{sql-r1}            &  \underline{87.6}          & -           & 66.6  & -    \\
    Reasoning-SQL + Qwen2.5-Coder-7B\citep{reasoning-sql}            & -          & -           & 64.0& -    \\

    \textbf{Ours: DB-Explore + Qwen2.5-Coder-7B}    & 87.2          & 80.2 & 65.2& 68.9\\
    \textbf{Ours: DB-Explore + Qwen2.5-Coder-7B + self-consistency@8}   & \textbf{87.8}& \textbf{81.9}& \textbf{67.0}& \textbf{71.2}\\
    \hline
  \end{tabular}
  \caption{\label{main}
    Experimental results on SPIDER and BIRD benchmarks. Asterisks '$^*$' denote performance re-evaluated through the official open-source implementation by ROUTE. Within the Fine-Tuning with Open-Source LLMs 7B group, top-performing entries are \textbf{bold} with runner-up scores \underline{underlined}. 
  }
\end{table*}

\paragraph{Benchmarks}  
(1) \textbf{SPIDER}  The SPIDER dataset \citep{yu2018spider} is a large-scale, cross-domain benchmark for Text-to-SQL tasks, designed to evaluate the generalization ability of models across diverse database structures. It consists of 8,659 training samples, and 1,034 development samples, spanning 200 databases and 138 domains.
(2) \textbf{BIRD}  The BIRD dataset \citep{liCanLLMAlready2023} is a more challenging Text-to-SQL benchmark compared to SPIDER, featuring 95 real-world databases across 37 professional domains, with a total storage size of 33.4GB.

\paragraph{Evaluation Metrics}  
In our experiments, we use three key evaluation metrics to assess the performance of Text-to-SQL models. Execution Accuracy (EX) \citep{yu2018spider} measures the proportion of queries where the predicted SQL query produces the exact output as the ground truth, serving as a direct indicator of correctness. Test-Suite Accuracy (TS) \citep{zhong2020semantic} stands out as a more trustworthy metric than EX. For BIRD, we report EX along with the Valid Efficiency Score (VES) \citep{liCanLLMAlready2023}, which is used to evaluate the execution efficiency of accurately generated SQL queries. Unlike EX, in VES the score for an accurately generated SQL query is no longer fixed at 1. Instead, it is determined by the ratio of the execution time of the ground truth SQL query to that of the predicted SQL query. 

\paragraph{Implementation Details}  
We employ GPT-4o \citep{hurst2024gpt} for data synthesis and fine-tune Qwen2.5-Coder-7B \citep{yang2024qwen2} as our base model. A total of 21,273 and 16,457 query-SQL pairs are generated for the BIRD and SPIDER datasets. All experiments are conducted on four NVIDIA A6000 GPUs with a batch size of 32. We utilize the Llama-Factory framework \citep{zheng2024llamafactory} for parameter-efficient fine-tuning. Both schema linking and text generation modules are trained for 3 epochs using AdamW optimization, with an initial learning rate of 1$e$-5 and a context window of 4,096 tokens. The learning rate follows a cosine decay schedule. For self-consistency decoding, we sample 8 SQL candidates with a temperature of 0.8. 

\paragraph{Baselines}  
We evaluate DB-Explore against state-of-the-art Text-to-SQL methods across two major paradigms. The first includes approaches \textbf{Prompting with Closed-Source LLMs}, such as MCS-SQL \citep{mcs} and Chase-SQL \citep{guoChaseLargeScalePragmatic2021}, which leverage multi-prompt strategies and result selection. MAC-SQL \citep{wangMACSQLMultiAgentCollaborative2023} and MAG-SQL \citep{mag} adopt multi-agent frameworks for error correction, while TA-SQL \citep{tasql}, DIN-SQL \citep{pourrezaDINSQLDecomposedInContext2023}, and DAIL-SQL \citep{dail} improve generation through task alignment, decomposition, and demonstration selection. The second group involves \textbf{Fine-Tuning Open-Source 7B LLMs}. DTS-SQL \citep{pourrezaDTSSQLDecomposedTexttoSQL2024} employs a two-stage pipeline, SENSE \citep{sense} leverages preference learning, and CODES \citep{liCodeSBuildingOpensource2024} incorporates extensive pretraining and plugins. ROUTE \citep{ROUTE} introduces multi-task learning, while SQL-o1 \citep{sql-o1}, SQL-R1 \citep{sql-r1}, and Reasoning-SQL \citep{reasoning-sql} integrate reinforcement learning to enhance SQL generation.


\section{Results and Analysis}

\subsection{Main Results}

\begin{figure}[h]
  \includegraphics[width=\columnwidth]{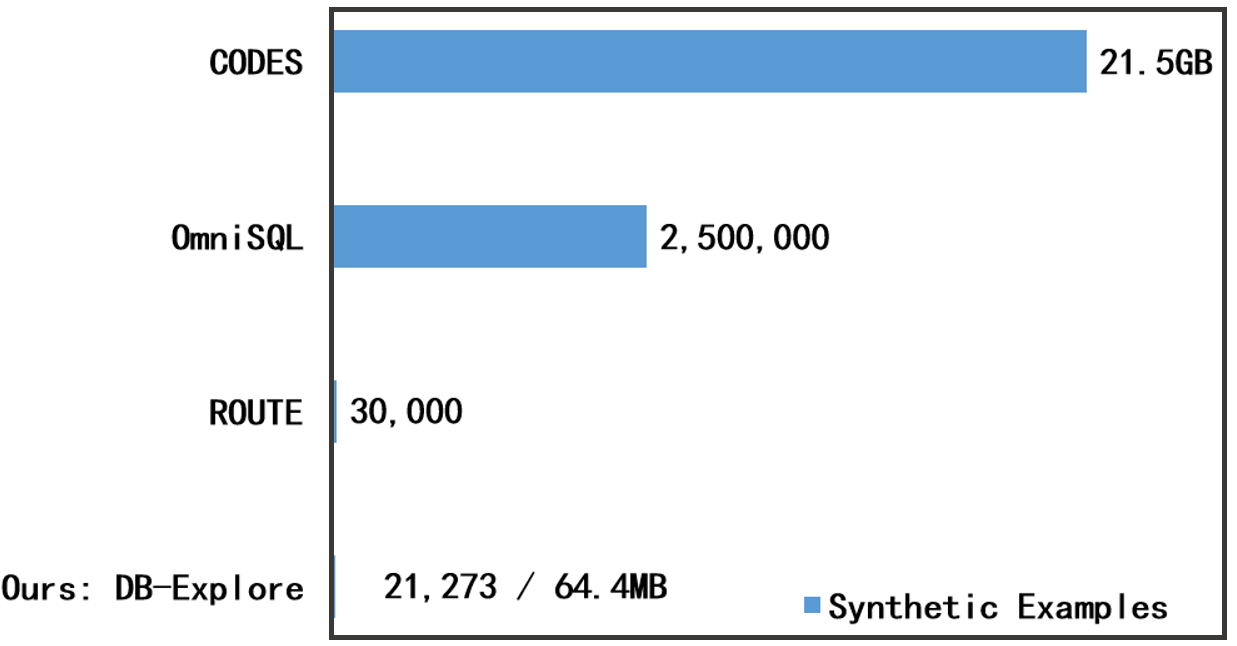}
  \caption{\label{fig4}Data volume analysis of different data synthesis methods on BIRD, CODES only reports the storage size for pre-training data}
\end{figure}

\begin{table*}[ht]
  \centering
  \begin{tabular}{lcccc}
    \hline
    ~                       & \multicolumn{2}{c}{SPIDER}    & \multicolumn{2}{c}{BIRD} \\
    ~                               & DB-Explore     & Base Model & DB-Explore     & Base Model \\
    \hline
    Full-Schema EX       & 83.9              & 80.0                         & 63.5           & 53.7    \\
    Filtered-Schema EX             & 87.2              & -                        & 65.2           & -    \\
    Gold-Schema EX             & 89.1             & 83.4                        & 69.9           & 63.8    \\
    \hline

  \end{tabular}
  \caption{\label{upper}
    The performance of the SQL generation model for different schema inputs, where the filtered-schema is the result of the DB-Explore schema linking model. 
  }
\end{table*}

\begin{table*}[ht]
  \centering
  \begin{tabular}{lcc}
    \hline
    Methods                       & {SPIDER}  & {BIRD}   \\
    ~                               & Dev-EX          & Dev-EX      \\
    \hline
    \textbf{Ours: DB-Explore + Qwen2.5-Coder-7B}    & \textbf{87.2}              & \textbf{65.2}\\

    - w/o Semantic Knowledge Extraction & 86.1$_{\downarrow1.1}$ & 64.8$_{\downarrow0.4}$ \\
    - w/o Structural Pattern Mining& 83.9$_{\downarrow3.3}$ & 62.5$_{\downarrow2.7}$     \\
    - w/o Progressive Instruction Synthesis & 84.8$_{\downarrow2.4}$ & 62.2$_{\downarrow3.0}$ \\
    - w/o Instruction Filtering  & 83.7$_{\downarrow3.5}$ & 61.4$_{\downarrow3.8}$ \\
    \hline
  \end{tabular}
  \caption{\label{ablation}
    Ablation study on DB-Explore, complete method of DB-Explore is \textbf{bold}
  }
\end{table*}

\paragraph{Performance on Main Benchmarks}
As shown in Table \ref{main}, DB-Explore achieves strong performance, attaining execution accuracies of 87.8\% on the SPIDER development set and 67.0\% on the BIRD development set—setting a new state-of-the-art among methods based on open-source 7B models. Compared to the baseline using Qwen2.5-Coder-7B, our approach yields substantial improvements. Notably, DB-Explore outperforms reinforcement learning-based methods such as SQL-R1 \citep{sql-r1} and Reasoning-SQL \citep{reasoning-sql}, despite relying only on limited supervised fine-tuning (SFT), significantly reducing training costs. Moreover, DB-Explore surpasses many Text-to-SQL methods that rely on closed-source models, including GPT-4 and GPT-4o. On the BIRD benchmark, it also achieves a VES score of 71.2\%, underscoring that our database‑specific exploration strategy can yield highly efficient query generation through capturing complex schema patterns.

\paragraph{Analysis of Data Volume}
Unlike existing approaches that focus on enhancing LLMs' generalization capabilities for SQL generation, DB-Explore adopts a database-centric data synthesis paradigm that prioritizes mining database-specific knowledge. This fundamental distinction enables our method to achieve competitive performance with significantly reduced data requirements. As illustrated in Figure \ref{fig4}, our approach achieves the lowest computational cost among all data synthesis-based Text-to-SQL methods. Specifically, DB-Explore demonstrates superior data efficiency on BIRD benchmark, achieving a significantly higher EX accuracy than other data synthesis methods while using only 70.9\% of the data used by ROUTE, and less than 1\% of that used by OmniSQL and CODES. 

\begin{figure}[h]
  \includegraphics[width=\columnwidth]{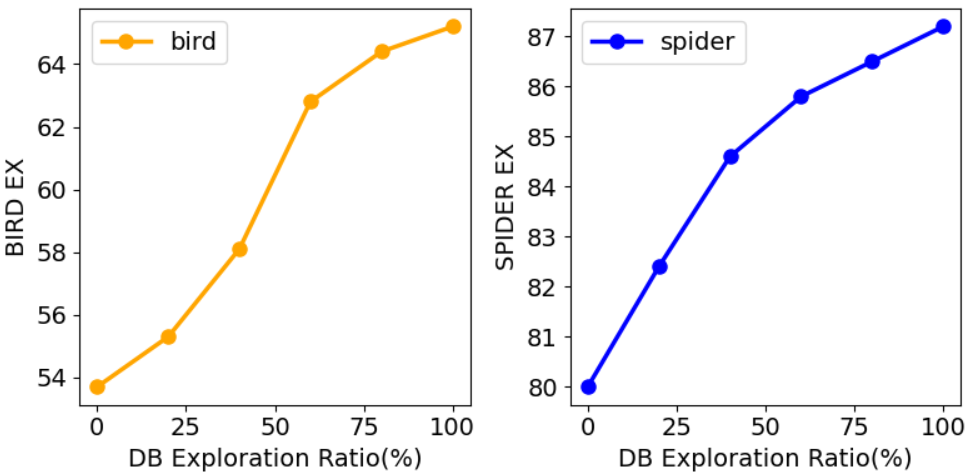}
\caption{\label{fig5}DB-Explore performance trend under different database exploration ratios.}
\end{figure}

\paragraph{Analysis of DB Exploration Ratio}
We analyze the impact of database exploration levels on model performance across the SPIDER and BIRD development sets. Different exploration levels are achieved by constraining the sampling scope of the DB Graph: 0\% corresponds to the Qwen2.5-Coder-7B baseline without exploration, while 100\% denotes full exploration as performed by DB-Explore. As shown in Figure \ref{fig5}, due to the relatively uniform distribution of queries across databases in both datasets, model performance increases almost linearly with the exploration ratio. Notably, database exploration not only enriches the model's schema understanding but also enhances its overall Text-to-SQL capabilities through fine-tuning. As a result, performance gains are slightly more pronounced in the early stages of exploration compared to the later stages.

\begin{figure}[h]
  \includegraphics[width=1\columnwidth]{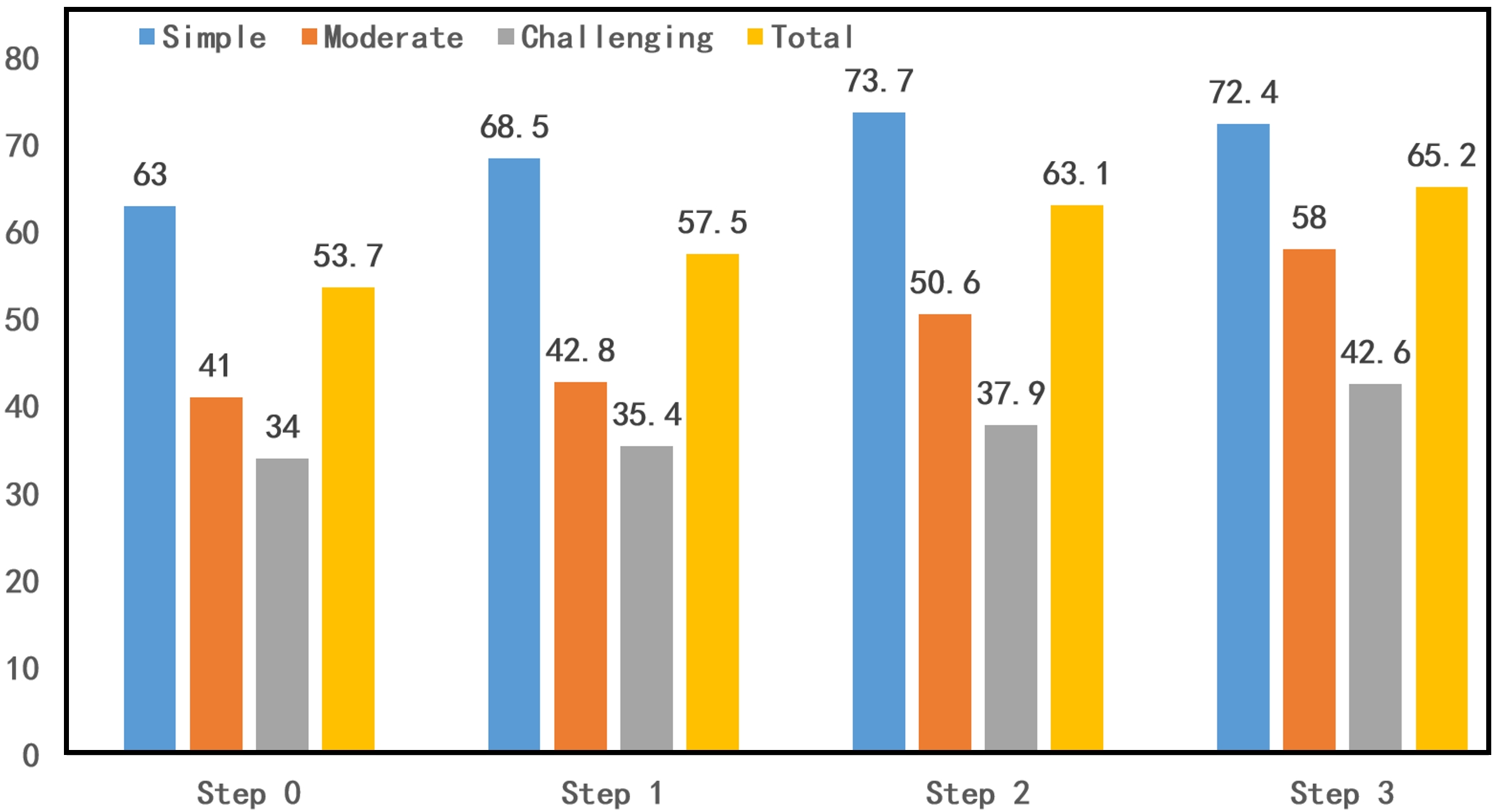}
\caption{\label{fig6}Performance comparison of DB-Explore at different instruction generation stages on BIRD development set.}
\end{figure}

\paragraph{Analysis of Progressive Instruction Synthesis}
To investigate the effect of progressive instruction generation on DB-Explore’s ability to handle complex queries, we analyze its performance across different instruction synthesis stages on the BIRD development set. Each step corresponds to one iteration of evolving the training instructions; to prevent excessive distributional drift, we retain all simpler instructions from prior steps at every iteration. As shown in Figure \ref{fig6}, DB-Explore exhibits a significant improvement in handling simple queries at Step 1, where the synthesized instructions remain relatively basic. However, performance on moderate and challenging queries shows little change at this stage. In Steps 2 and 3, as the synthesized data becomes increasingly complex through iterative evolution, the model's performance on moderate and challenging queries improves rapidly. As iterations accumulate, the synthesized instructions at Step 3, which now encompass queries involving four tables and five tables, are substantially more complex than those in the evaluation set. Although we retain all simpler instructions from Step 1 and Step 2, the overall data distribution at Step 3 still shifts slightly toward the more difficult examples, leading to a modest decline in performance on simple queries.

\paragraph{Analysis of Schema Linking}
As shown in Table \ref{upper}, the schema linking module not only alleviates the context length limitations of LLMs but also effectively focuses their attention on relevant schema elements. Incorporating Filtered-Schema leads to a 3.3\% and 1.7\% execution accuracy improvement on the SPIDER and BIRD datasets. Moreover, DB-Explore achieves strong performance when provided with the gold schema, further demonstrating its effectiveness on the Text-to-SQL task.

\subsection{Ablation Study}

\textbf{Study on Semantic Knowledge Extraction and Structural Pattern Mining}
We conducted ablation experiments to quantify the individual and joint effects of Semantic Knowledge Extraction and Structural Pattern Mining on DB‑Explore’s performance. As shown in Table \ref{ablation}, when the Semantic Knowledge Extraction module is removed, accuracy decreases by 1.1\% on SPIDER and 0.4\% on BIRD, indicating its role in extracting entity categories and predicate relationships that guide precise column and value mapping. By contrast, disabling Structural Pattern Mining which captures foreign‑key dependencies and the global topology of the database schema results in a much larger 3.3\% drop on SPIDER and 2.7\% drop on BIRD, primarily due to increased join‑path selection failures and ambiguous column predictions. Due to the inherent semantic information present in the synthesized data, Structural Pattern Mining brings even larger gains than Semantic Knowledge Extraction.

\textbf{Study on Progressive Instruction Synthesis}
To assess the impact of gradually increasing query complexity, we evaluated DB‑Explore with and without the Progressive Instruction Synthesis module. Incorporation of this module yields a 2.4\% improvement on SPIDER and a 3.0\% gain on BIRD, with the majority of benefits observed on moderate and challenging queries that require multi‑table joins and nested filters.

\textbf{Study on Instruction Filtering}
Finally, we examined the necessity of our Instruction Filtering component, which prunes hallucinated or incomplete instructions produced during synthesis. Without this filter, performance on SPIDER declines by 3.5\% and on BIRD by 3.8\%, reflecting the disruptive influence of low‑quality examples. This ablation confirms that rigorous filtering is indispensable for maintaining instruction quality and preventing performance degradation caused by spurious or malformed queries. 


\section{Conclusion}
In this paper, we propose DB-Explore, a framework for database exploration, instruction synthesis, and fine-tuning that enhances the LLMs' understanding of databases by extracting dynamic database knowledge, thereby unlocking the potential of LLMs in Text-to-SQL tasks. Our approach, through database multi-knowledge exploration, effectively minimizes the risks associated with insufficient database knowledge in SQL generation. We demonstrate the efficacy and superiority of our method on recent LLMs across several benchmark tests. The results show that our approach achieves state-of-the-art performance in Text-to-SQL with minimal computational cost. In the future, we aim to explore more database knowledge, larger LLMs, and more efficient fine-tuning frameworks to achieve robust and efficient Text-to-SQL solutions. 

\section*{Limitations}

While our exploration demonstrates promising results, several important limitations warrant acknowledgment. First, the framework introduces modest computational cost and GPT-4o API consumption. Although these costs remain substantially lower than comparable methods, we encourage future work to explore more efficient data generation and fine-tuning paradigms. Second, our upper-bound analysis reveals a persistent performance gap between our approach and inference frameworks directly leveraging closed-source models. Finally, due to computational constraints, the framework's effectiveness on larger language models (e.g., Qwen2-72B \citep{hui2024qwen2}) remains unverified, leaving its scalability to state-of-the-art architectures as open research questions.

\bibliography{main}

\clearpage

\end{document}